\documentclass{article}
\usepackage{ijcai22}

\usepackage{times}
\usepackage{soul}
\usepackage{url}
\usepackage[hidelinks]{hyperref}
\usepackage[utf8]{inputenc}
\usepackage[small]{caption}
\usepackage{graphicx}
\usepackage{amsmath,amssymb}
\usepackage{amsthm}
\usepackage{booktabs}
\usepackage{multirow}
\usepackage{algorithm}
\newtheorem{definition}{Definition}

\title{SGAT: Simplicial Graph Attention Network}

\author{
See Hian Lee
\and
Feng Ji\And
Wee Peng Tay
\affiliations
Nanyang Technological University, Singapore
\emails
seehian001@e.ntu.edu.sg, \{jifeng, wptay\}@ntu.edu.sg
}

\usepackage{mathtools}
\usepackage{amsthm}
\usepackage{scalerel}
\usepackage{nicefrac}
\usepackage[shortlabels]{enumitem}
\usepackage{graphicx}
\usepackage{epstopdf}
\DeclareGraphicsExtensions{.eps,.png,.jpg,.pdf}

\usepackage{url}
\usepackage{colortbl}
\usepackage{booktabs}
\usepackage{multirow}
\usepackage[normalem]{ulem}
\usepackage{xparse}
\usepackage{calc}
\usepackage{etoolbox}
\usepackage{amsfonts}

\usepackage[capitalize]{cleveref}
\crefname{equation}{}{}
\Crefname{equation}{}{}
\crefname{claim}{claim}{claims}
\crefname{step}{step}{steps}
\crefname{line}{line}{lines}
\crefname{condition}{condition}{conditions}
\crefname{dmath}{}{}
\crefname{dseries}{}{}
\crefname{dgroup}{}{}

\crefname{Problem}{Problem}{Problems}
\crefformat{Problem}{Problem~(#2#1#3)}
\crefrangeformat{Problem}{Problems~(#3#1#4) to~(#5#2#6)}

\crefname{Theorem}{Theorem}{Theorems}
\crefname{Corollary}{Corollary}{Corollaries}
\crefname{Proposition}{Proposition}{Propositions}
\crefname{Lemma}{Lemma}{Lemmas}
\crefname{Definition}{Definition}{Definitions}
\crefname{Example}{Example}{Examples}
\crefname{Assumption}{Assumption}{Assumptions}
\crefname{Remark}{Remark}{Remarks}
\crefname{Rem}{Remark}{Remarks}
\crefname{remarks}{Remarks}{Remarks}
\crefname{Appendix}{Appendix}{Appendices}
\crefname{Supplement}{Supplement}{Supplements}
\crefname{Exercise}{Exercise}{Exercises}
\crefname{Theorem_A}{Theorem}{Theorems}
\crefname{Corollary_A}{Corollary}{Corollaries}
\crefname{Proposition_A}{Proposition}{Propositions}
\crefname{Lemma_A}{Lemma}{Lemmas}
\crefname{Definition_A}{Definition}{Definitions}

\newcommand{\ifbcdot}[1]{\ifblank{#1}{\cdot}{#1}}

\DeclarePairedDelimiterX\abs[1]{\lvert}{\rvert}{\ifbcdot{#1}}
\DeclarePairedDelimiterX\parens[1]{(}{)}{\ifbcdot{#1}}
\DeclarePairedDelimiterX\brk[1]{[}{]}{\ifbcdot{#1}}
\DeclarePairedDelimiterX\braces[1]{\{}{\}}{\ifbcdot{#1}}
\DeclarePairedDelimiterX\angles[1]{\langle}{\rangle}{\ifblank{#1}{\cdot,\cdot}{#1}}
\DeclarePairedDelimiterX\ip[2]{\langle}{\rangle}{\ifbcdot{#1},\ifbcdot{#2}}
\DeclarePairedDelimiterX\norm[1]{\lVert}{\rVert}{\ifbcdot{#1}}
\DeclarePairedDelimiterX\ceil[1]{\lceil}{\rceil}{\ifbcdot{#1}}
\DeclarePairedDelimiterX\floor[1]{\lfloor}{\rfloor}{\ifbcdot{#1}}

\DeclarePairedDelimiterXPP\trace[1]{\operatorname{Tr}}{(}{)}{}{\ifbcdot{#1}} 
\DeclarePairedDelimiterXPP\col[1]{\operatorname{col}}{\{}{\}}{}{\ifbcdot{#1}} 
\DeclarePairedDelimiterXPP\row[1]{\operatorname{row}}{\{}{\}}{}{\ifbcdot{#1}} 
\DeclarePairedDelimiterXPP\erf[1]{\operatorname{erf}}{(}{)}{}{\ifbcdot{#1}}
\DeclarePairedDelimiterXPP\erfc[1]{\operatorname{erfc}}{(}{)}{}{\ifbcdot{#1}}
\DeclarePairedDelimiterXPP\KLD[2]{D}{(}{)}{}{\ifbcdot{#1}\, \delimsize\|\, \ifbcdot{#2}} 
\DeclarePairedDelimiterXPP\op[2]{\operatorname{#1}}{(}{)}{}{#2} 

\DeclarePairedDelimiterXPP\indicate[1]{{\bf 1}}{\{}{\}}{}{\ifbcdot{#1}}

\newcommand{\ofrac}[1]{{\frac{1}{#1}}}

\newcommand{\tc}[1]{^{(#1)}}

\providecommand\given{}
\newcommand\SetSymbol[2][]{%
	\nonscript\, #1#2
	\allowbreak
	\nonscript\,
	\mathopen{}}

\DeclarePairedDelimiterX\Set[2]\{\}{%
\renewcommand\given{\SetSymbol[\delimsize]{#1}}
#2
}
\DeclarePairedDelimiterX\Setc[1]\{\}{%
\renewcommand\given{\SetSymbol{:}}
#1
}

\NewDocumentCommand\set{s o m}{%
	\IfBooleanTF#1%
	{\IfValueTF{#2}{\Set*{#2}{#3}}{\Setc*{#3}}}%
	{\IfValueTF{#2}{\Set{#2}{#3}}{\Setc{#3}}}%
}

\DeclareMathOperator*{\concat}{\scalerel*{\parallel}{\sum}}

\newcommand{\msout}[1]{\text{\color{green} \sout{\ensuremath{#1}}}}
\newcommand{\del}[1]{{\color{green}\ifmmode \msout{#1}\else\sout{#1}\fi}}

\usepackage[noend]{algpseudocode}
\algnewcommand{\LineComment}[1]{\State \(\triangleright\) #1}

\begin{document}

\maketitle

\begin{abstract}
Heterogeneous graphs have multiple node and edge types and are semantically richer than homogeneous graphs. To learn such complex semantics, many graph neural network approaches for heterogeneous graphs use metapaths to capture multi-hop interactions between nodes. Typically, features from non-target nodes are not incorporated into the learning procedure. However, there can be nonlinear, high-order interactions involving multiple nodes or edges. In this paper, we present Simplicial Graph Attention Network (SGAT), a simplicial complex approach to represent such high-order interactions by placing features from non-target nodes on the simplices. We then use attention mechanisms and upper adjacencies to generate representations. We empirically demonstrate the efficacy of our approach with node classification tasks on heterogeneous graph datasets and further show SGAT's ability in extracting structural information by employing random node features. Numerical experiments indicate that SGAT performs better than other current state-of-the-art heterogeneous graph learning methods.
\end{abstract}

\section{Introduction}

Real world objects are often connected and interdependent, with graph-structured data collected in many applications. Recently, graph neural network (GNN), which learns tasks utilizing graph-structured data, has attracted much attention. Initial models such as GraphSage \cite{hamilton2017graphsage}, Graph Convolution Network (GCN) \cite{kipf2016semi} and Graph Attention Network (GAT) \cite{velivckovic2017graph} focused on homogeneous graphs in which all nodes and edges are of the same type. Works to extend these models to heterogeneous graphs, which have multiple node or edge types, have then been proposed \cite{yun2019gtn} as the majority of real-world graphs are heterogeneous in nature. These models typically employ metapaths \cite{fu2020magnn}.

A metapath is a predefined sequence of node types describing a multi-hop relationship between nodes \cite{dong2017metapath2vec}. For instance, in a citation network with author, paper and venue node types, the metapath Author-Paper-Author~(APA) represents co-author relationship. Moreover, Author-Paper-Venue-Paper-Author (APVPA) connects two authors who published papers at the same venue. To generalize further, the concept of a metagraph is introduced \cite{mhin_metagraph}. A metagraph is a nonlinear combination of metapaths or more specifically, a subgraph of node types defined on the network schema, describing a more elaborated pairwise relation between two nodes through auxiliary nodes. We can therefore view metapaths and metagraphs as multi-hop relations between two nodes.

Although metapath-based methods can achieve state of the art performances, they have at least one of the following limitations: (1) The model is sensitive to metapaths with suboptimal paths leading to suboptimal results \cite{just2018}. Metapath selection requires domain knowledge of the dataset and the task at hand. Likewise, metagraphs are predefined based on domain knowledge. Alternatively, heuristic algorithms can be employed to extract the most common structures to form metagraphs \cite{mhin_metagraph}. Yet, the generated metagraphs may not necessarily be useful \cite{metagraph-spectral2018}. (2) Metapath-based models do not incorporate features from non-target nodes along the metapaths, thus discarding potentially useful information \cite{fu2020magnn}. Examples include HERec \cite{HERec} and Heterogeneous graph Attention Network (HAN) \cite{han2019}. (3) The model relies on \emph{structures between two nodes} to capture complex pairwise relations and semantics. This, however, is not always sufficient to capture more complex interactions in graphs which simultaneously involve multiple target nodes and cannot be reduced to pairwise relationships, especially for heterogeneous graphs given their richer semantics \cite{bunch2020simplicial}.


Several works have aimed to improve the expressive power of GNNs and generate more accurate representations of heterogeneous graph components without utilizing metapaths or metagraphs. One potential alternative is a simplicial complex. Simplicial complexes are natural extensions of graphs that can represent high-order interactions between nodes. A graph defines pairwise relationships between elements of a vertex set. Meanwhile, a simplicial complex defines higher order relations, e.g., a 3-tuple is a triangle, a 4-tuple is a tetrahedron and so on (for a proper definition, see \cref{paragraph:SC}). For instance, a triangle made up of three author vertices in a citation network indicates co-authorship of the same paper between the three authors, whereas in a graph, edges between pairs of these three authors only tell us that they are pairwise co-authors with each other on some papers. These structures are already employed in Topological Data analysis (TDA) to extract information from data and in other applications such as tumor progression analysis \cite{tumoranalysis} and brain network analysis \cite{brainnetworkanalysis} to represent complex interactions between elements. The GNN literature \cite{bunch2020simplicial,mpsn_icml21} on simplicial complexes thus far have focused on homogeneous graphs. These works cannot be directly applied to heterogeneous graphs.

In this paper, we propose a general framework, SGAT, which extends GAT to heterogeneous graphs with simplicial complexes. SGAT learns on heterogeneous graphs using simplicial complexes to model higher order relations and passing messages between the higher-order simplices. We first describe a procedure to generate $k$-order homogeneous simplices from a heterogeneous graph since heterogeneous datasets do not always possess higher-order simplices, given their innate schemas. In order to avoid discarding potentially useful information when transforming the heterogeneous graph into homogeneous simplices, we populate the $k$-simplices, for $k\geq1$, with non-target node features and learn the importance of each of the $k$-order simplicies through attention mechanisms with upper adjacencies. Overall, the contributions of this paper are as follows:
\begin{itemize}
  \item We develop a procedure to construct a simplicial complex from a heterogeneous graph. Our proposed procedure converts the graph into a homogeneous simplicial complex without loss of feature information.
  \item We propose GAT-like attention mechanisms that operate on simplicial complexes. We utilize upper adjacencies to pass messages between higher-order simplices to learn effective embeddings that capture higher order interactions. We also introduce a variant model SGAT-EF that incorporates edge features.
  \item We apply SGAT to the node classification task on standard heterogeneous graph datasets, which demonstrate that our proposed approach outperforms current state-of-the-art models. We additionally assess the ability of our model in extracting structural information using random node features.
\end{itemize}

\section{Related Work}\label{sec:related}
In this section, models that are designed for heterogeneous graphs and related to SGAT are reviewed. Learning on heterogeneous graphs mainly utilizes predefined metapaths or automatically learning the optimal metapaths during the end-to-end training process \cite{just2018}. Works along this line include HAN, HERec, metapath2vec \cite{dong2017metapath2vec}, Graph Transformer Network (GTN) \cite{yun2019gtn} and REGATHER \cite{regather}. 

HAN utilizes predefined metapaths to transform heterogeneous graphs into metapath-based homogeneous graphs, discarding features of non-target nodes. After which, dual-level attention is applied to the metapath-based homogeneous graphs. On the other hand, HERec employs metapaths to generate metapath-based random walks, then filters the walks to contain only the node type of the starting node. This avoids representing nodes of different node types in the same unified space. Similarly, metapath2vec utilizes metapaths to guide random walks to incorporate semantic relations and avoid bias towards more visible node types. Given the difficulty in selecting optimal metapaths, GTN proposes to learn metapath-based graph structures by softly selecting candidate adjacencies and multiplying them together. The resulting weighted adjacency matrices represent metapath-based graph structures. Meanwhile, REGATHER generates a set of multi-hop relation-type subgraphs, which indirectly includes metapath-based graph structures by multiplying adjacency matrices of first order relation-types during data preprocessing. Attention mechanism is then utilized to assign different weights to the subgraphs. All of these approaches do not learn high-order relations in the TDA context \cite{mhin_metagraph}. 

Metagraphs have been proposed to express finer-grained, non-linear semantics. Approaches that utilize metagraphs are M-HIN \cite{mhin_metagraph}, mg2vec \cite{mg2vec} and Meta-GNN \cite{meta-gnn-asonam2019}. M-HIN constructs triplets to illustrate the relationship between nodes and metagraphs and subsequently, applies the Hadamard function in complex space to encode the relationship between nodes and metagraphs. By utilizing metagraphs and a complex embedding scheme, M-HIN is able to capture more accurate node features. As for mg2vec, embeddings of metagraphs and nodes are simultaneously learned and mapped into a common low-dimensional space. Metagraphs are utilized to guide the learning of node representations and to capture the latent relationships between nodes. Lastly, Meta-GNN introduces a metagraph convolution layer, employing metagraphs to define the receptive field of nodes. Attention mechanism is also used to combine node features from different metagraphs.

Homogeneous GNN approaches involving simplicial complexes include Message Passing Simplicial Network (MPSN) \cite{mpsn_icml21} and Simplicial Neural Network (SNN) \cite{ebli2020simplicial}. MPSN introduces a general message passing framework on simplicial complexes, describing four different adjacencies that simplices can have. Besides that, they also introduced the Simplicial Weisfeiler-Lehman (SWL) test and showed that the SWL test is strictly more powerful than the WL test \cite{douglas2011weisfeilerlehman} in distinguishing non-isomorphic graphs. Since standard GNNs with local neighborhood aggregation are shown to be equivalent to the WL test in their expressive powers, complementing GNNs with simplicial complexes can help increase their expressive powers. Nevertheless, MPSN is specific to homogeneous graphs and the application of simplicial complexes to heterogeneous graphs is unexplored by \cite{mpsn_icml21}.

SNN is a framework utilizing simplicial complexes on heterogeneous bipartite graphs. Heterogeneous bipartite graphs are transformed into a homogeneous complex with corresponding co-chains' data. The co-chain features are then refined by spectral simplicial convolution with Hodge Laplacians \cite{tsp_barbarossa} of the simplicial complex. Although SNN also populates features on high-order structures, it is limited to bipartite graphs and its objective is to impute missing data. In addition, SNN operates on the spectral domain, which is different from our proposed approach that operates on the graph domain directly. To the best of our knowledge, our proposed framework is the first study to learn heterogeneous graphs (which are not limited to bipartite graphs) with simplicial complexes.

\section{Simplicial Graph Attention Network}\label{sec:SGAT}

We first introduce the concept of a simplicial complex, its adjacencies, and related notations. The reader is referred to the \cref{appendix:SC-intro} and \cite{Hatcher:478079} for further details.

\paragraph{Simplicial Complexes.}\label{paragraph:SC} A simplicial complex $\chi$ is a collection of finite sets (simplices) closed under subsets. A member of $\chi$ is a called a $k$-simplex and denoted as $\sigma^k$ if it has cardinality $k+1$. A $k$-simplex has $k+1$ faces of dimension $k-1$. A face is obtained by omitting one element from the simplex. Specifically, by omitting the $j$-th element, a face of a $k$-simplex is a set containing elements of the form $(v_0,\ldots,v_{j-1},v_{j+1},\ldots,v_{k})$. Subsequently, it is possible to think of 0-simplices $\sigma^0$ as vertices, 1-simplices $\sigma^1$ as edges, 2-simplices $\sigma^2$ as triangles, 3-simplices $\sigma^3$ as tetrahedron and so on. Lastly, the dimension of $\chi$ is the largest dimension of any simplex in it and is denoted by $K$. The set of all simplices in $\chi$ of dimension $k$ is denoted by $\chi^k$ with cardinality $|\chi^k|$. 

\paragraph{Simplicial Adjacencies.} In this paper, we utilize the notion of upper-adjacency to denote the neighborhood of simplices. Two $k$-simplices in $\chi^k$ are upper-adjacent if they are faces of the same $(k+1)$-simplex. For instance, two edges are upper-adjacent if they are part of a same triangle. Hence, the neighborhood of a $k$-simplex, $N(\sigma^k_i)$ is the set of $k$-simplices upper-adjacent to it. We include self-loops to the upper adjacency matrix so that a $k$-simplex is upper-adjacent to itself. We have
\begin{align}
\label{eqn:neighborhood}
N(\sigma^k_i) = \set*{\sigma^k_j \in \set*{\sigma^k_1 \ldots \sigma^k_{|\chi^k|}} \given A^{k}(i,j)=1},
\end{align}
where $A^k \in \mathbb{R}^{|\chi^k| \times |\chi^k|}$ denotes the upper adjacency matrix indicating whether pairs of $k$-simplices are upper-adjacent.


\subsection{Construction of Simplices}\label{subsec:SC-construct}
Many real-world heterogeneous graph datasets do not have predefined higher-order $k$-simplices for $k\geq2$ due to their unique schemas. We propose a procedure to transform heterogeneous graphs into homogeneous simplices where the 0-simplices are the singleton subsets of the target node type's vertices. The target node type is the node type whose labels we want to predict (node classification). Meanwhile, non-target nodes features reside on higher order simplices.

Consider a heterogeneous graph $G=(V,E)$ with $a$ edge-types and $b$ node types, where $a+b>2$. The set of nodes is $V$ and the set of edges is given by $E$. Let $\tilde{V}\subset V$ be the set of target nodes. We assume that each $v\in\tilde{V}$ has a feature vector $h_v$. To simplify the presentation, we assume that every non-target node $u\in V\backslash\tilde{V}$ is also associated with a feature vector $h_u$ (if a non-target node does not come with a feature vector, we can associate with it a one-hot encoding of its node type).

For $1\leq k \leq K$, where $K$ is a hyperparameter, we choose the $k$-simplices to be the sets of $k+1$ target vertices that share at least $\epsilon^{k}_{\eta}$ common non-target neighbors which are exactly $\eta$ hops away, where $\eta>0$ is a chosen parameter. This is performed for each $\eta\leq\eta_{\max}$, where $\eta_{\max}$ is a hyperparameter, to generate $\eta$ number of $K$ dimensional simplicial complexes. Consider a $k$-simplex $\sigma^k$, where $k>1$, whose vertices share $m \geq \epsilon^{k}_{\eta}$ common $\eta$-hop non-target neighbors, ${s_1, \ldots, s_m}$. The $k$-simplex is assigned the feature $h_{\sigma^k}=\phi({h_{s_1}, \ldots, h_{s_m}})$ where $\phi$ is the average operator.
For 1-simplices, the features of the intermediate node(s) along a path between the 0-simplices are first summed to form the path feature, $\Theta_{\sigma^1}=\sum_{u\in\theta}{h_{u}}$ where $\theta$ is the set of intermediate nodes along the path connecting the faces of $\sigma^1$. The path features between the same pair of 0-simplices are then averaged and assigned as the feature of the 1-simplex. In addition, we place the simplex's own feature on the connecting simplex when a self-loop is present.

\begin{figure}[!htb]
\centering
\includegraphics[width=0.9\columnwidth]{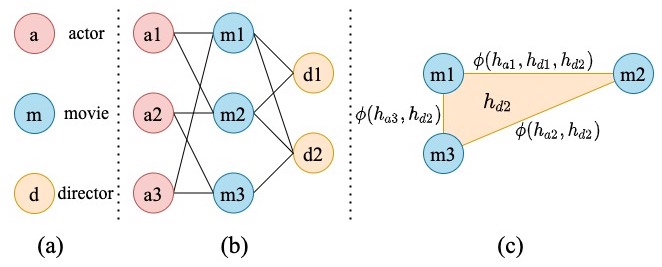}
\caption{Constructing simplicial complexes from the heterogeneous IMDB graph (a)-(c) with $\eta=1$. 
(a) IMDB node types. (b) IMDB graph illustration. (c) Resulting movie one-hop sharing complex with $\epsilon_1^1=1$.
}\label{fig:simplicial}
\end{figure}

Taking the IMDB\footnote{https://www.imdb.com/interfaces/} dataset as an example in \cref{fig:simplicial}, we set $\eta=1$ and $\epsilon^1_1=1$. The constructed simplicial complex is illustrated in \cref{fig:simplicial}. We note that the set of 1-simplices is different from the original edges in the graph.  

The technical details to construct all the $k$-simplices of a single simplicial complex within a network is given in \cref{alg:pseudocode} in \cref{appendix:pseudocode}

\subsubsection{Incorporation of Edge Features} In the case where each edge $(i,j)\in E$ has a feature vector $e_{ij}$, we concatenate the feature generated as described above for a $1$-simplex $h_{\sigma^1} = \phi({\Theta_{\sigma^1,1}}, 
\ldots, {\Theta_{\sigma^1,t}})$ where $t$ is the number of paths between the faces of $\sigma^1$, with the average of the edge features along the original path in the graph that gives rise to $\sigma^1$. For $\sigma^1$ with end nodes $\tilde{v}$ and $v$, we have
\begin{align}
h_{\sigma^1} &=\phi({\Theta_{\sigma^1,1}}, 
\ldots, {\Theta_{\sigma^1, t}}) \parallel \phi(\Phi_1, \ldots, \Phi_m),\\
\Phi_i &= \phi(e_{\tilde{v} s_{i,1}}, \ldots, e_{s_{i,j}s_{i,j+1}}, \ldots, e_{s_{i,\kappa} v}),
\end{align}
where $\kappa$ is the number of intermediate nodes between the end nodes and $\Phi_i$ is the summarized edge feature along each of the original $m$ paths. We note that many heterogeneous GNNs such as GTN and HAN are not designed to incorporate edge features during their learning process.

\begin{figure}[!htb]
\centering
\includegraphics[width=0.65\columnwidth]{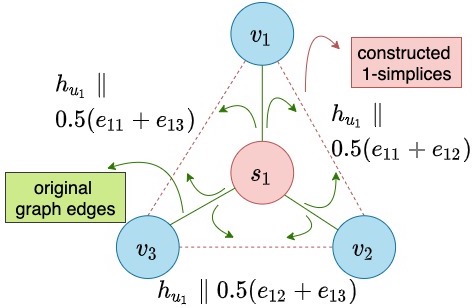}
\caption{Combining edge features with features of 1-simplices to enrich features of 1-simplices for $\eta=1$.
}\label{fig:edge_ft}
\end{figure}

When $n>k$ target nodes share more than $\epsilon_{\eta}^k$ non-target nodes among them, a total of ${\binom{n}{k+1}}$ $k$-simplices are constructed, which may cause memory overloading issues for large values of $n$.
To further control the the amount of constructed simplices, a hyperparameter $\lambda > K$ is introduced. We do not construct the simplices (and their corresponding faces) if $n\geq\lambda$. This can also be interpreted as disregarding the $k$-simplices (and their associated faces) where $k\geq\lambda$.

\begin{figure*}[!htb]
\centering
	\includegraphics[width=0.97\textwidth]{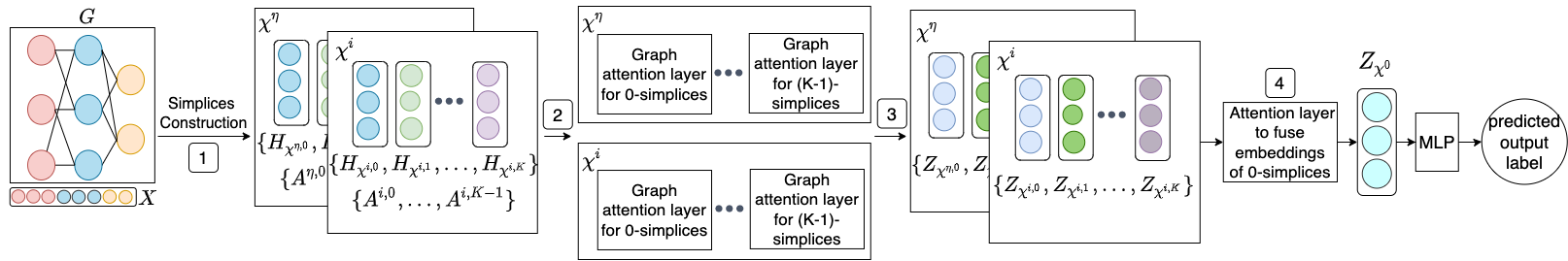}
\caption{The overall architecture of SGAT (self-loops are omitted for clarity). (1) Given the heterogeneous graph $G$ and node features $X$ as inputs, we first construct $K$ simplices for each of the $\eta$ simplicial complex(es) and the simplices' respective features. For each simplicial complex $\chi^i$, (2) its associated simplices, features and upper adjacency matrices are fed into a simplicial attention layer which is made up of $K$ graph attention layers. Then, (3) $\eta$ sets of simplicial complex specific embeddings are generated. For node classification task, (4) we fuse the $\eta$ sets of $0$-simplices' embeddings using attention to obtain the final embedding which is then fed into a layer of MLP (the classifier).
}\label{fig:architecture}
\end{figure*}

\subsection{Simplicial Attention Layer}\label{subsec:SAL}

After constructing the $k$-simplices, $0\leq k \leq K$, from the original graph, the inputs to the SGAT model are $K+1$ different sets of features, $\set*{H_{\chi^0}, H_{\chi^1}, \ldots, H_{\chi^K}}$, where $H_{\chi^k}=\set*{h_{\sigma^k_1}, \ldots, h_{\sigma^k_{|\chi^k|}}}$, $h_{\sigma^k_i}$ is the feature vector of the $i$-th $k$-simplex, $i = 1,\ldots, |\chi^k|$. Moreover, each of the $K+1$ feature sets is associated with an upper adjacency matrix, $A^k$.

We employ a simplicial graph attention layer to compute the importance of the neighboring simplices and update the various simplices' features. A simplicial graph attention layer is made up of $K$ graph attention layers, one for each of the $k$-order simplices, $0\leq k < K$, excluding the $K$-order. This captures the interactions between higher-order structures of the same order by passing messages between themselves through upper adjacencies. For each of the $K$ graph attention layers, given a $k$-simplex pair $\sigma_i^k, \sigma_j^k$, the unnormalized attention score between the two simplicies is defined as

\begin{equation}
\resizebox{\linewidth}{!}{$
    \displaystyle
    \zeta_{ij}^{k} =\mathrm{LeakyReLU}\parens*{({\bf a}^k)^{\intercal} \brk*{{\bf W}^k h_{\sigma^{k}_i}\, \delimsize\|\, {\bf W}^k h_{\sigma^{k}_j} \, \delimsize\|\, {\bf W}^{k+1} h_{\sigma^{k+1}_{ij}}}},
$}
\end{equation}%

where $\parallel$ represents concatenation, $^{\intercal}$ denotes transposition, ${\bf a}^k$ is the learnable attention vector and ${\bf W}^k$ is the weight matrix for a linear transformation specific to the $k$-order and are shared by all the $k$-simplices. The feature vector $h_{\sigma^{k+1}_{ij}}$ belongs to the $k+1$-simplex that both $\sigma_{i}^k$ and $\sigma_{j}^k$ are faces of.

The normalized attention score for each of the $k$-simplices can then be obtained by applying the softmax function as given by:
\begin{align}
\alpha_{ij}^{k} = \frac{\exp(\zeta_{ij}^{k})}{\sum_{r\in N(\sigma^k_i)}\exp(\zeta_{ir}^{k})},
\end{align}
where $N(\sigma^k_i)$ refers to the neighborhood of $\sigma_i^k$ (cf.\ \cref{eqn:neighborhood}). The weight coefficients learned can then be applied along with the corresponding neighbors' features to update the feature for each of the $k$-simplices as given by
\begin{align}\label{eq:zsigma}
z_{\sigma^{k}_i} = \mu\parens*{\sum_{j\in N(\sigma^k_i)} \alpha_{ij}^k \mathbf{W}^k h_{\sigma^{k}_j}},
\end{align}
where $\mu(\cdot)$ denotes an activation function. To stabilise the training, we employ multi-head attention where $P$ independent attention mechanisms are performed. The results are then concatenated to generate the required learned simplex features:
\begin{align}
z_{\sigma^{k}_i} = \concat_{p=1}^{P} \mu\parens*{\sum_{j\in N(\sigma^k_i)} \alpha_{ij}^{k,p} \mathbf{W}^{k,p} h_{\sigma^{k}_j}},
\end{align}
where $\alpha_{ij}^{k,p}$ and $\mathbf{W}^{k,p}$ are the attention and transformation weights corresponding to $\alpha_{ij}^k$ and $\mathbf{W}^k$, respectively, in \cref{eq:zsigma} for the $p$-th attention head.

The output for the $k$-order simplices $\chi^k$ is a set of simplicial complex specific embeddings, $Z_{\chi^k} = \set{z_{\sigma^k_1},\ldots,z_{\sigma^k_{|\chi^k|}}}$. Since we have $K$ graph attention layers, there are $K$ sets of simplicial complex specific embeddings $Z_{\chi^0},\ldots, Z_{\chi^{K-1}}$. Depending on the downstream task, not all of the embeddings are utilized. For instance, in the last softmax layer for node classification tasks, only the $Z_{\chi^0}$ embeddings are employed.

\subsection{Attention to Fuse Simplicial Complexes}

Recall that for $\eta_{\max}>1$ and for each $\eta=1,\ldots,\eta_{\max}$, we construct $k$-simplices for $0\leq k \leq K$. Applying the simplicial attention layer in \cref{subsec:SAL} to each of the $\eta$ number of $K$ dimensional simplicial complexes, we obtain the embeddings $Z_{\chi^{0,\eta}}, \ldots, Z_{\chi^{K-1,\eta}}$ for each $\eta$-specific simplicial complex.
In our framework, a $k$-simplex $\sigma^k$ is a unique $k$-tuple that is independent of $\eta$. We arbitrarily order all $k$-simplices as $\sigma^k_1, \sigma^k_2, \ldots, \sigma^k_\tau$, where $\tau=\max_\eta |\chi^{k,\eta}|$. The collection of embeddings for $\chi^{k,\eta}$ can then be written as $Z_{\chi^{k,\eta}}=\set{z^{\eta}_{\sigma^{k}_{1}}, \ldots, z^{\eta}_{\sigma^{k}_{\tau}}}$ where $z^{\eta}_{\sigma^{k}_i}$ is the feature vector of the $i$-th $k$-simplex for the $\eta$-specific simplicial complex, which is taken to be null if $\sigma^k_i \notin \chi^{k,\eta}$. 

In this subsection, we propose an attention layer to account for the importance of different simplicial complexes in describing the respective $k$-dimensional simplices. An attention layer is required to fuse the $\eta$ groups of $k$-simplex embeddings. 

The unnormalized attention score to fuse the $\eta$-specific embeddings of $k$-simplices is defined as follows:
\begin{equation}\label{equation:attn_fusion}
\resizebox{\linewidth}{!}{$
    \displaystyle
    w^{k,\eta} = \ofrac{|\chi^{k,\eta}|}\sum_{i=1}^{\tau} (\mathbf{q}^{k, \eta})^{\intercal} \tanh(\mathbf{F}^{k, \eta} z^{\eta}_{\sigma^{k}_i} +\mathbf{b}^{k, \eta})\indicate*{\sigma^k_i \in \chi^{k,\eta}},
$}
\end{equation}%
where $\mathbf{q}^{k, \eta}$ is the learnable attention vector, $\mathbf{F}^{k, \eta}$ is a learnable weight matrix and $\mathbf{b}^{k, \eta}$ is a learnable bias vector. Here, $\indicate{A}$ is the indicator function, which takes value $1$ if $A$ is true and $0$ otherwise. The normalized weight coefficients can then be attained through the softmax function as
\begin{align}
\beta^{k,\eta} = \frac{\exp(w^{k,\eta})}{\sum_{j=1}^{\eta_{\max}} \exp(w^{k,j})}.
\end{align}

We fuse the features $z^{\eta}_{\sigma^{k}_{i}}$ across $\eta=1,\ldots,\eta_{\max}$ by a linear combination using the learned weights to generate the final embedding of $\sigma^k_i$ as
\begin{align}
z_{\sigma^k_i} = \sum_{\eta=1}^{\eta_{\max}} \beta^{k,\eta} z^\eta_{\sigma^k_i}\indicate*{\sigma^k_i\in\chi^{k,\eta}}.
\end{align}
Let $Z_{\chi^{k}}=\set*{z_{\sigma^{k}_{1}}, \ldots, z_{\sigma^{k}_{\tau}}}$ be the embedding of $\chi^k=\set{\sigma^k_1,\ldots,\sigma^k_\tau}$.

Lastly, when $L$ layers are utilized, let $Z\tc{l}_{\chi^0}$ be the embedding of $\chi^0$ in the $l$-th layer, for $l=1,\ldots,L$. We concatenate the learned 0-simplex features of each layer at the last layer to obtain
\begin{align}
Z_{\chi^{0}}^{(L)} = \concat_{l=1}^{L} Z_{\chi^{0}}^{(l)},
\end{align}
and feed it into a linear layer for our semi-supervised, node classification task along with cross-entropy loss. Nonetheless, we note that the final learned features, $Z_{\chi^{0}}^{(L)}$ can be used for other downstream tasks similarly optimising the model via backpropagation. The overall architecture of SGAT is as shown in \cref{fig:architecture}.

\begin{table*}[!htb]
\centering
\resizebox{\linewidth}{!}{
\begin{tabular}{@{}cccccccccc@{}}

\toprule
Datasets                    & Metrics  & GCN              & GAT            & Meta-GNN          & HAN                  & REGATHER            & GTN                  & SGAT                  & SGAT-EF        \\ \midrule
\multirow{2}{*}{DBLP}       & Macro-F1 & 87.65 $\pm$ 0.29 & 91.69 $\pm$ 0.27 & 92.47 $\pm$ 0.92  & 91.93 $\pm$ 0.27     & 91.79 $\pm$ 0.69    & 93.59 $\pm$ 0.40     & \bf{93.80 $\pm$ 0.20} & 93.73 $\pm$ 0.23 \\
                            & Micro-F1 & 88.71 $\pm$ 2.74 & 92.65 $\pm$ 0.25 & 93.48 $\pm$ 0.75  & 92.51 $\pm$ 0.24     & 92.70 $\pm$ 0.67    & 94.17 $\pm$ 0.26     & \bf{94.58 $\pm$0.20}  & 94.51 $\pm$ 0.19 \\ \hline
\multirow{2}{*}{ACM}        & Macro-F1 & 91.46 $\pm$ 0.48 & 92.16 $\pm$ 0.32 & 88.74 $\pm$ 0.99  & 91.01 $\pm$ 0.76     & 92.38 $\pm$ 0.57    & 92.23 $\pm$ 0.60     & 92.41 $\pm$ 0.36      & \bf{92.91 $\pm$ 1.79} \\
                            & Micro-F1 & 91.33 $\pm$ 0.47 & 92.06 $\pm$ 0.33 & 88.71 $\pm$ 1.01  & 90.93 $\pm$ 0.73     & 92.30 $\pm$ 0.56    & 92.12 $\pm$ 0.62     & 92.35 $\pm$ 0.36      & \bf{92.86 $\pm$ 1.75} \\ \hline
\multirow{2}{*}{IMDB}       & Macro-F1 & 56.72 $\pm$ 0.49 & 57.32 $\pm$ 0.88 & 56.19 $\pm$ 0.97  & 56.56 $\pm$ 0.77     & 56.34 $\pm$ 0.54    & 59.12 $\pm$ 1.58     & 59.97 $\pm$ 0.41      & \bf{60.36 $\pm$ 0.53} \\
                            & Micro-F1 & 58.31 $\pm$ 0.51 & 58.75 $\pm$ 0.98 & 58.68 $\pm$ 1.49  & 57.83 $\pm$ 0.93     & 57.59 $\pm$ 0.64    & 60.58 $\pm$ 2.10     & 62.51 $\pm$ 0.64      & \bf{62.74 $\pm$ 0.77} \\ \bottomrule
\end{tabular}}
\captionsetup{justification=centering}
\caption{Node classification result on heterogeneous datasets (Standard split). Averaged over five runs, best performance boldfaced.}
\label{table:cls_result2}
\end{table*}

\section{Numerical Experiments}\label{sec:experiments}
In this section, we verify the empirical performance of our proposed method against several state-of-the-art methods on node classification task for heterogenous graphs.

\subsection{Datasets}
The heterogeneous datasets utilized are two citation network datasets DBLP\footnote{https://dblp.uni-trier.de/} and ACM, and a movie dataset IMDB. Since these heterogeneous datasets do not consist of edge features, when edge features are required, we form $h_e$ for each $e \in E$ of the respective datasets by concatenating its starting node feature, ending node feature and a one-hot encoding of its edge-type. The dataset statistics are in \cref{appendix:dataset}.



\subsection{Baselines and Settings} \label{sec:baselines}
We compare SGAT and SGAT-EF with six state-of-the-art GNN models. For all models, the hidden units are set to 64, the Adam optimizer was used and its hyperparameters such as learning rate and weight decay, are respectively chosen to yield best performance. When metapaths are required, the metapaths employed in \cite{han2019} are utilized. As for metagraphs, the relevant metapaths and 3-node metagraphs as in \cite{meta-gnn-asonam2019} are used.

For SGAT, we set $K$, the dimension of the simplicial complexes to be 2, the number of layers to be 2 for all the datasets. Moreover, when $\eta>=2$, the dimension of the attention vector $\mathbf{q}^{k, \eta}$ (cf.\ \cref{equation:attn_fusion}) is set to 128. Besides the parameters mentioned above, $\epsilon^{k}_{\eta}$, $\eta$ and $\lambda$ are tuned for each dataset. Specifically, for ACM, we choose $\eta=1$, $\epsilon^{1}_{1}=1$, and $\lambda=20$. For DBLP, $\eta=2$, $\epsilon^{1}_{1}=3$, $\epsilon^{1}_{2}=4$, and $\lambda=10$. For IMDB, $\eta=1$, $\epsilon^{1}_{1}=1$, and $\lambda=10$.

\subsection{Node Classification Performance}
\cref{table:cls_result2} shows the performances of SGAT, SGAT-EF and other node classification baselines. Our empirical results demonstrate that SGAT and SGAT-EF outperform the baselines for node classification task. We observe that SGAT performs better than GTN, HAN and REGATHER. This demonstrates that nonlinear structures encoded by the simplicial complexes are useful for learning more effective node representations and more complicated relationships, which may not be possible using linear structures such as metapaths. 

\section{Model Analysis}
\subsection{Ablation Study}
We conduct experiments on different SGAT variants to validate the effectiveness of the components of our model. SGAT considers message passing between higher-order simplices without incorporating edge features while SGAT-EF is the equivalent model utilizing edge features. Moreover, since SGAT generalizes to the GAT model when $K=1$, the GAT model can also be taken as an ablation study of SGAT without message passing on higher order structures. As observed in \cref{table:cls_result2}, by employing message passing on higher-order simplices, SGAT obtained a significant improvement over GAT. We also observe that SGAT-EF frequently performs better than SGAT. This show that our method incorporating edge features may bring benefits when the task of interest is edge dependent or when the dataset has meaningful edge features. 

A parameter sensitivity study (for $\lambda$ and $\epsilon^k_\eta$) is given in \cref{appendix:parame_sensitivity}

\subsection{Meta-GNN, SGAT and SGAT-EF}
\begin{figure}[!htb]
\centering
	\includegraphics[width=0.99\columnwidth]{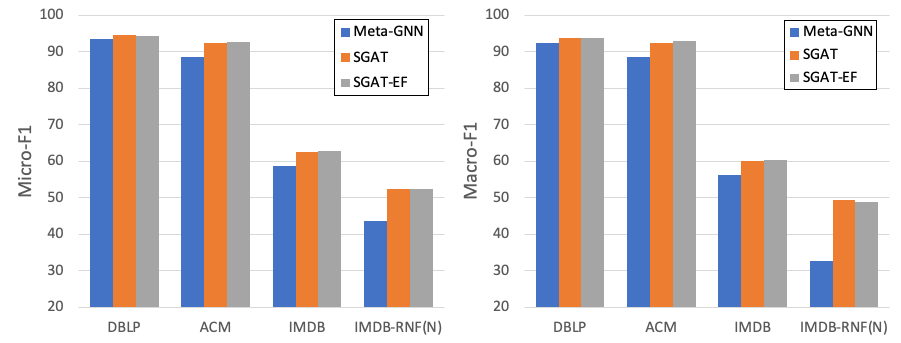}
\caption{Node classification F1 scores (\%) for Meta-GNN, SGAT and SGAT-EF
}\label{fig:meta_vs_sgat}
\end{figure}

To evaluate the advantage of utilising simplicial complexes over metagraphs, we conduct experiments to compare Meta-GNN, SGAT and SGAT-EF. The results are shown in \cref{fig:meta_vs_sgat}. We can clearly observe that SGAT and SGAT-EF, which performs message passing between simplices, consistently outperforms Meta-GNN. Meta-GNN uses metagraphs, which despite being a subgraph pattern, solely represents relationships between \emph{two} target nodes and not \emph{multiple} target nodes as in the case of simplices. The result demonstrates that metagraphs cannot adequately represent high-order interactions between multiple target nodes. 

\subsection{Random Node Features}

A survey \cite{graphkernelsurvey} has shown that node features of benchmark graphs usually contain substantial information, allowing models to achieve close to optimal results. Hence, we prepare an additional dataset to classify nodes solely based upon graph structure. We replace node features of IMDB dataset by random ones sampled from standard normal distribution, naming it IMDB-RNF(N). The random node features are the same size as the replaced node features. This leaves only the graph structure for the classification task and thus, we can utilise the resulting performance to interpret the models' capability in extracting structural information. This approach is also used in \cite{horn2021topological}. Note that for IMDB-RNF(N) dataset, the models are observed to be highly sensitive, thus we re-tuned all the models to ensure a fair comparison. \cref{table:cls_result_rnf} depicts the results on such dataset.

\begin{table}[!htb]
    \centering
    \begin{tabular}{lcl}
    \toprule
             & \multicolumn{2}{c}{IMDB-RNF(N)}                     \\
    Methods  & Macro-F1                           & Micro-F1                \\ \midrule
    GAT      & 36.45 $\pm$ 2.03                     & 37.94 $\pm$ 2.07         \\
    GCN      & 40.10 $\pm$ 1.09                     & 41.84 $\pm$ 1.24          \\
    Meta-GNN & 32.58 $\pm$ 1.62                     & 43.75 $\pm$ 5.62          \\
    HAN      & 39.47 $\pm$ 1.03                     & 40.97 $\pm$ 1.13          \\
    REGATHER & 28.60 $\pm$ 3.45                     & 42.62 $\pm$ 8.10          \\
    GTN      & 33.71 $\pm$ 0.51                     & 36.52 $\pm$ 1.38          \\ \midrule
    SGAT     & \textbf{49.29 $\pm$ 1.34}            & 52.34 $\pm$ 1.81          \\
    SGAT-EF  & \multicolumn{1}{l}{48.90 $\pm$ 1.00} & \textbf{52.37 $\pm$ 1.09} \\ \bottomrule
    \end{tabular}
    \captionsetup{justification=centering}
    \caption{Node classification result on heterogeneous dataset with random node features, best performance boldfaced.}
    \label{table:cls_result_rnf}
    \end{table}



We observe clear advantage of SGAT and SGAT-EF over their comparison partners. For instance, SGAT and SGAT-EF surpass the baselines by up to approximately 10\% F1 scores on the IMDB-RNF(N) dataset. In addition, we find that the performance of all models decreased when the bag-of-words node features are replaced with uninformative ones. This shows that the original node features indeed contain useful information for the task, aiding all models to garner almost optimal results.  Subsequently, it also implies that our models are able to effectively extract and utilise structural information from graphs instead of relying on the original node features that may already contain important information.

\section{Conclusion}\label{sec:conclusion}
In this paper, we introduced SGAT, a novel extension of GAT to heterogeneous graphs using simplicial complexes and upper adjacencies. SGAT does not utilize metapaths or metagraphs. Hence, SGAT is not subjected to the three characteristic limitations of metapath-based methods, namely (1) performance being sensitive to the choice of metapaths, (2) discarding non-target node features and (3) limited expressiveness when using structures between two nodes to capture complex interactions. Specifically, we utilize simplicial complex to capture higher order relations among multiple target nodes. We avoid discarding non-target node features when transforming the graph into homogeneous simplicies by placing those features on simplices. We also introduced a variant incorporating edge features that can boost embedding performance. Empirically, we demonstrated that SGAT performs favorably on node classification task for heterogeneous datasets and even when random node features was employed.  



\appendix
\section{Dataset Statistics} \label{appendix:dataset}
We perform node classification task on three heterogeneous benchmark datasets. Characteristics of the datasets are summarised in \cref{table:dataset_stats}. DBLP has three node types (author(A), paper(P) and conference(C)), and the research area of author serve as labels. ACM consists of three node types (paper(P), author(A) and subject(S)), and categories of papers are to be predicted. IMDB has three node types (movie(M), actor(A), and director(D)). The labels to be determined are the genres of the movies. Each node in DBLP, ACM and IMDB has an associated, bag-of-words node feature.

\begin{table}[!htb]
\centering
\resizebox{\columnwidth}{!}{
\begin{tabular}{cccccc} 
\toprule
\multicolumn{1}{c}{Dataset} & \multicolumn{1}{c}{\# Nodes} & \multicolumn{1}{c}{\# Edges} & {\# Node type} &{\# Classes}& {\# Features} \\ 
\midrule
DBLP                        & 18405                        & 67946                        & 3                &  4            & 334\\
ACM                         & 8994                         & 25922                         & 3               &  3            & 1902\\
IMDB                        & 12772                        & 37288                        & 3                &  3            & 1256\\
\bottomrule
\end{tabular}}
\captionsetup{justification=centering}
\caption{Summary of datasets}
\label{table:dataset_stats}
\end{table}

\section{Pseudocode} \label{appendix:pseudocode}
For simplicity, we assume that the $\epsilon^k_{\eta}$ is the same for all $k$-order simplices and the specified $\eta$. The Geometric Understanding in Higher Dimensions library \cite{gudhi} for computational topology is utilised to ensure the constructed simplices form simplicial complex(es) that adhere to the inclusion property.

\begin{algorithm}[tb]
    \caption{Construction of $k$-simplices}\label{alg:pseudocode}
    \textbf{Input}: The adjacency list of heterogeneous graph Adj\_list,\\
    Node features $X$,\\
    Number of shared non-target neighbours $\epsilon$,\\
    Number of hops away $\eta$,\\
    Maximal $k$-order considered, $K$,\\
    The maximum simplex order to construct $\lambda$.\\
    \textbf{Output}: Set of all $k$-simplices, AllKSimplices.
    
    \begin{algorithmic}[1]
    \LineComment{$\texttt{DFSFindPaths}$ is a modified depth first search that breaks at $2\eta$ depth. To get paths that starts and ends with target node type. Each path should contain at least one non-target node.}
    \LineComment{Each path is of form [target\_src, node\_k(s), target\_dst], where node\_k is in $\theta$ and $\theta$ is the set of intermediate nodes along the path.}
    \State Initialise PathList as empty list.
    \For{$v \in V$}
            \State PathList $\gets \texttt{DFSFindPaths}$(Adj\_list$, v, 2\eta)$ 
        \EndFor
    
    \LineComment{The middle node refers to the non-target node $\eta$ hops away from the target\_src and target\_dst.}
    \State Initialise MidNodeNeighborDict, $D_{mid}$ \LineComment{where key is MidNodeID and value is its set of unique Neighbors.}
    \For{path $\in$ PathList}
        \State $D_{mid}$[path.MidNodeID].insert(path.src)
        \State $D_{mid}$[path.MidNodeID].insert(path.dst)
    \EndFor
    
    \State Initialise KSimplicesDict 
    \LineComment{where key is the Neighbors and value is the list of MidNodeID}
    \For{MidNodeID, Neighbors $\in D_{mid}$}
        \If {$2 \leq $ size(Neighbors) $\leq \lambda$}
            \State KSimplicesDict[Neighbors].insert(MidNodeID)
        \EndIf
    \EndFor
    
    \For{Neighbors, MidNodeIDList in KSimplicesDict}
        \If{size(MidNodeIDList) $\geq \epsilon$}
            \LineComment{Simplex tree is a data structure in Gudhi library}
            \State SimplexTree.insert(Neighbors)
        \EndIf
    \EndFor
    \LineComment{Get up to $K$ simplices from SimplexTree}
    \State AllKSimplices $\gets$ SimplexTree.get\_simplices()
    \State \Return AllKSimplices
    \end{algorithmic}
    \end{algorithm}
    

\section{Relation to Hypergraphs}
In this paper, the primary geometric concept is that of simplicial complexes, as a generalization of graphs. A simplicial complex is a special class of hypergraph. We choose to work exclusively with simplicial complexes instead hypergraphs for a few reasons. A simplicial complex has rich algebraic structure and plays a central role in algebraic topology. Many tools have been developed to study simplicial complexes which are unavailable for general hypergraphs, such as the Hodge-de Rham theory. Moreover, for any given hypergraph, we may construct a simplicial complex by treating each hyperedge as a simplex and including hyperedges associated with its faces. The resulting simplicial complex has many important geometric properties identical to the initial hypergraph. In this respect, using simplicial complexes is sufficient to extract useful geometric information from the datasets.

\section{Parameter Sensitivity} \label{appendix:parame_sensitivity}
\begin{figure}[!htb]
\centering
	\includegraphics[width=1\columnwidth]{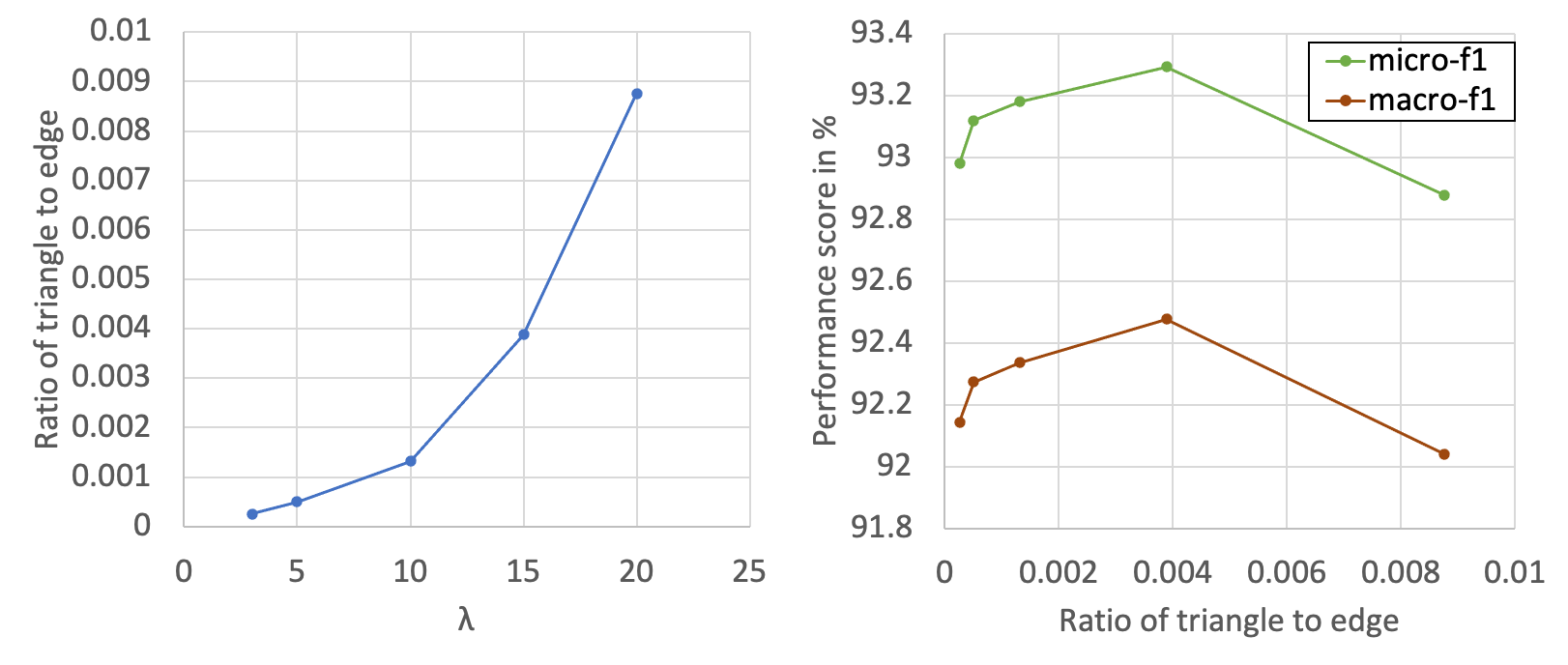}
\caption{$\lambda$ against $\gamma$ (left) and $\gamma$ against performance score (right)}\label{chart:lambda}
\end{figure}

Given that SGAT involves a number of parameters to control the construction of simplices, we further examine how $\lambda$ (the maximum simplex order to construct, including their faces. The simplex order refers to its dimension) and $\epsilon^{k}_{\eta}$ (the minimum number of shared non-target nodes that is $k$ and $\eta$-specific) affect the performance of SGAT on the DBLP dataset. We first keep all the constructed edges by setting $\epsilon^{1}_{1}=\epsilon^{1}_{2}=1$ and measure the ratio of constructed triangles to constructed edges, $\gamma$ as a function of $\lambda$. The Micro-F1 and Macro-F1 scores are then obtained as a function of $\gamma$ as seen in \cref{chart:lambda}. We observe that an optimal $\gamma$ where the information included from the amount of simplices considered is most beneficial and does not negatively affect the model's performance. Increasing $\lambda$ beyond the resulting optimal $\gamma$ is likely to introduce noise by including unnecessary messages being passed between the simplices.

We also analyse how $\epsilon^{k}_{\eta}$ affect SGAT's performance by setting $\lambda$ to be fixed at 10 and the default values of $\epsilon^{1}_{1}=3$ and $\epsilon^{1}_{2}=4$. Besides the parameter being tested, the other parameters assume their default values. When $\epsilon^{1}_{1}$ and $\epsilon^{1}_{2}$ are increased, the number of constructed edges decreases. We observe that increasing $\epsilon^{1}_{1}$ improves the performance until the default value of $\epsilon^{1}_{1}=3$ is reached. After which, the performance deteriorates. This phenomenon is similar for $\epsilon^{1}_{2}$. Increasing $\epsilon^{k}_{\eta}$ reduces noise in the model as it removes weaker edges (those that shared fewer nodes).

\begin{figure}[!htb]
\centering
	\includegraphics[width=1\columnwidth]{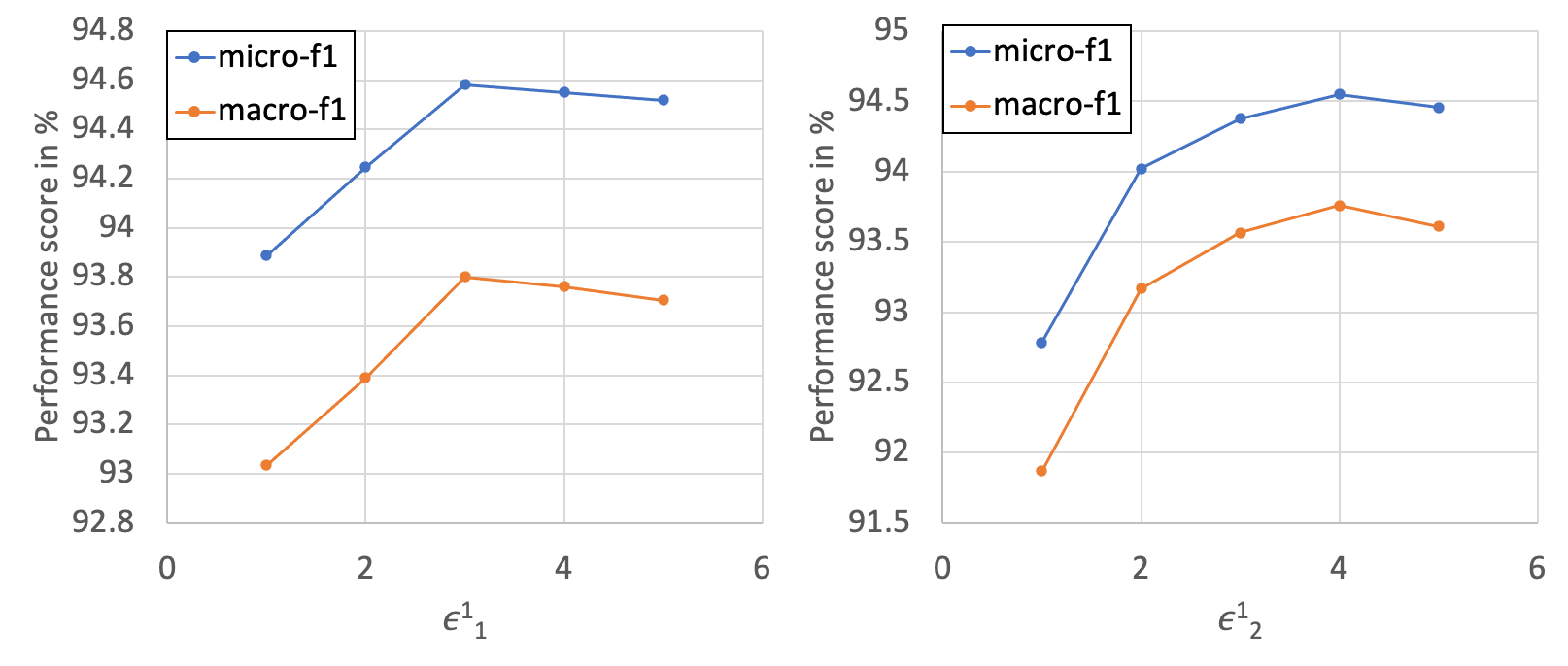}
\caption{$\epsilon^{1}_{1}$ against $\gamma$ (left) and $\epsilon^{1}_{2}$ against $\gamma$ (right)}\label{chart:epsilon}
\end{figure}

\begin{figure}[!htb]
\centering
	\includegraphics[width=1\columnwidth]{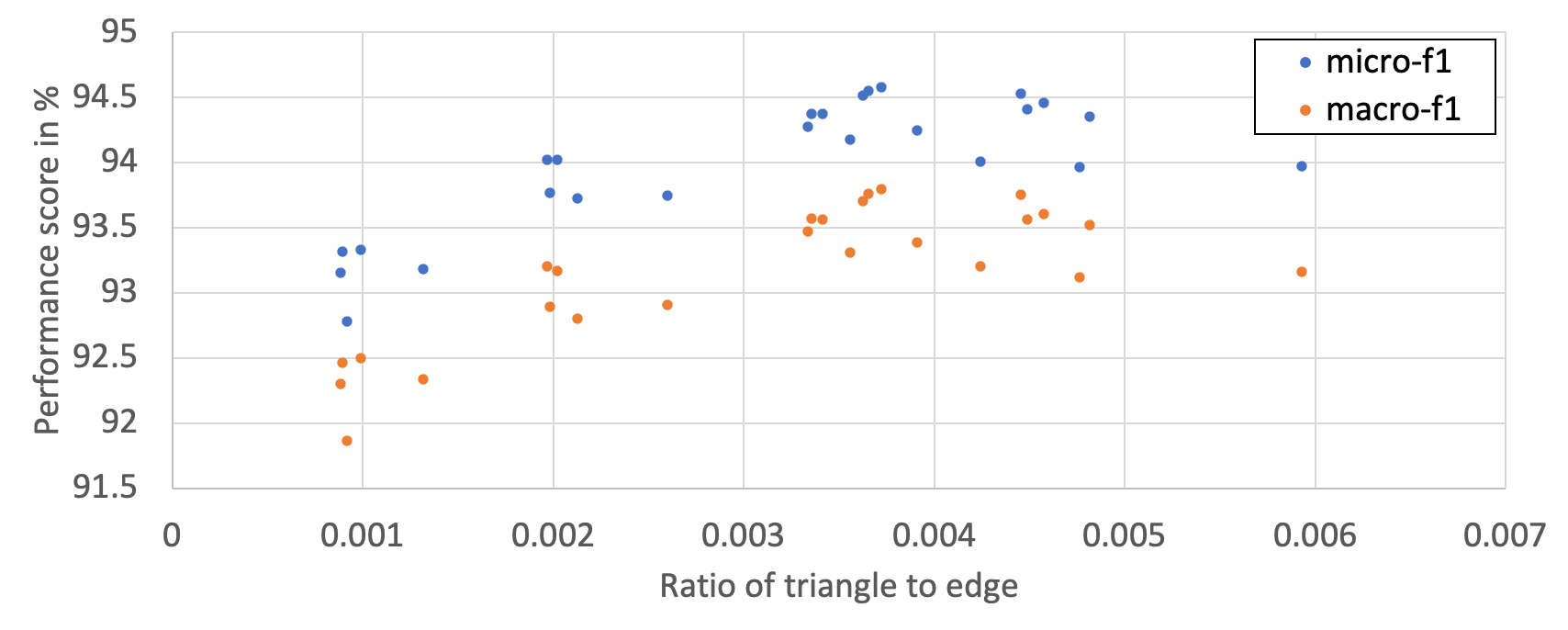}
\caption{$\gamma$ against performance score}\label{chart:combination}
\end{figure}

Lastly, we examine how different combinations of $\epsilon^{1}_{1}$ and $\epsilon^{1}_{2}$, each chosen in the range of $[1,5]$, affect SGAT's performance. Interestingly, we found that the performance of SGAT fluctuates minimally across different values of these parameters when the ratio of triangles to edges, $\gamma$ is similar. 

\section{Preliminaries}\label{appendix:SC-intro}
In this section, we give a brief overview of simplicial complexes. We refer interested readers to \cite{Hatcher:478079} for more details. Here, we discuss some properties of simplicial complexes and the notion of upper adjacency, which we use in our proposed SGAT model.

\subsection{Simplicial Complex}\label{subsec:SC}
A simplicial complex is a set consisting of vertices, edges and other higher order counterparts, all of which are known as $k$-simplices, $k\geq0$, defined as follows.

\begin{definition}[k-simplex]\label{def:k-simplex}
A standard (unit) $k$-simplex is defined as 
\begin{align}\label{sigmak}
\sigma^{k} = \set*{(\tilde{v}_0,\ldots,\tilde{v}_{k}) \in \mathbb{R}^{k+1}_{\geq 0} \given \sum_{i=0}^{k}{\tilde{v}_i = 1}}.
\end{align}
Any topological space that is homeomorphic to the standard $k$-simplex and thus, share the same topological characteristics is called a $k$-simplex. 
\end{definition}
As an example, let ${v}_0,\ldots,{v}_{k}$ be vertices of a graph embedded in a Euclidean space so that they are affinely independent (i.e., $v_1-v_0, \ldots, v_k-v_0$ are linearly independent). Then the set 
\begin{align*}
\set*{\sum_{i=0}^k \alpha_i v_i \given \alpha_i\geq 0,\ \sum_{i=0}^k\alpha_i =1}
\end{align*}
 is a $k$-simplex. In particular,the vertices $\{v_i\}$ are instances of $\sigma^0$ while the edges $(v_i, v_j)$, $i\ne j$, are instances of $\sigma^1$. Subsequently, it is possible to geometrically interpret 0-simplices $\sigma^0$ as vertices, 1-simplices $\sigma^1$ as edges, 2-simplices $\sigma^2$ as triangles, 3-simplices $\sigma^3$ as tetrahedron and so on. 

A face of a $k$-simplex is a $(k-1)$-simplex that we obtain from \cref{sigmak} by restricting a fixed coordinate in $(\tilde{v}_0,\ldots,\tilde{v}_{k})$ to be zero. Specifically, a face of a $k$-simplex is a set containing elements of the form $(\tilde{v}_0,\ldots,\tilde{v}_{j-1},\tilde{v}_{j+1},\ldots,\tilde{v}_{k})$. A $k$-simplex has $k+1$ faces.

A simplicial complex is a class of topological spaces that encodes higher-order relationships between vertices. The formal definition is as below.
\begin{definition}[Simplicial complex]
A simplicial complex $\chi$ is a finite set of simplices such that the following holds.
\begin{itemize}
  \item Every face of a simplex in $\chi$ is also in $\chi$.
  \item The non-empty intersection of any two simplices $\sigma_1, \sigma_2$ in $\chi$ is a face of both $\sigma_1$ and $\sigma_2$.
\end{itemize}
\end{definition}

It is also possible to produce a concrete geometric object for each simplicial complex $\chi$. The geometric realisation of a simplicial complex is a topological space formed by glueing simplices together along their common faces \cite{ji2020signal}. For instance, a simplicial complex of dimension 1, consists of two kinds of simplices $\sigma^0$ and $\sigma^1$. By glueing $\sigma^1$ with common $\sigma^0$, we obtain a graph in the usual sense. This means that $\chi^0$ is the set of vertices $V$ in the graph and $\chi^1$ is the set of edges $E$ in the graph.


\subsection{Simplicial Adjacencies}
Four kinds of adjacencies can be identified between simplices. They are boundary adjacencies, co-boundary adjacencies, lower-adjacencies and upper-adjacencies  \cite{tsp_barbarossa}. In this paper, we utilize only upper-adjacency so that our SGAT model is equivalent to GAT when $K=1$.
We say that two $k$-simplices in $\chi^k$ are upper-adjacent if they are faces of the same $(k+1)$-simplex. The upper adjacency matrix $A^k \in \mathbb{R}^{|\chi^k| \times |\chi^k|}$ of $\chi^k=\set{\sigma^k_1 \ldots \sigma^k_{|\chi^k|}}$ indicates whether pairs of $k$-simplices are upper-adjacent. The upper adjacency matrix of 0-simplices (nodes), $A^0$, is the usual adjacency matrix of a graph.

\section*{Acknowledgements}
The first author is supported by Shopee Singapore Private Limited under the Economic Development Board Industrial Postgraduate Programme (EDB IPP). The programme is a collaboration between Shopee and Nanyang Technological University, Singapore. The last two authors are supported in part by the Singapore Ministry of Education Academic Research Fund Tier 2 grant MOE-T2EP20220-0002.
\bibliographystyle{named}
\bibliography{ijcai22}

\begin{thebibliography}{}

\bibitem[\protect\citeauthoryear{Barbarossa and
  Sardellitti}{2020}]{tsp_barbarossa}
Sergio Barbarossa and Stefania Sardellitti.
\newblock Topological signal processing over simplicial complexes.
\newblock {\em IEEE Trans.\ on Signal Processing}, PP:1--1, 03 2020.

\bibitem[\protect\citeauthoryear{Bodnar \bgroup \em et al.\egroup
  }{2021}]{mpsn_icml21}
Cristian Bodnar, Fabrizio Frasca, Yuguang Wang, Nina Otter, Guido~F Montufar,
  Pietro Li{\'o}, and Michael Bronstein.
\newblock Weisfeiler and lehman go topological: Message passing simplicial
  networks.
\newblock In {\em Proc.\ of the 38th International Conference on Machine
  Learning (ICML)}, pages 1026--1037, 2021.

\bibitem[\protect\citeauthoryear{Borgwardt \bgroup \em et al.\egroup
  }{2020}]{graphkernelsurvey}
Karsten Borgwardt, Elisabetta Ghisu, Felipe Llinares-López, Leslie O’Bray,
  and Bastian Rieck.
\newblock {\em Graph Kernels: State-of-the-Art and Future Challenges},
  volume~13.
\newblock Foundations and Trends® in Machine Learning, 2020.

\bibitem[\protect\citeauthoryear{Bunch \bgroup \em et al.\egroup
  }{2020}]{bunch2020simplicial}
Eric Bunch, Qian You, Glenn Fung, and Vikas Singh.
\newblock Simplicial 2-complex convolutional neural networks.
\newblock {\em NeurIPS 2020 Workshop on Topological Data Analysis and Beyond},
  2020.

\bibitem[\protect\citeauthoryear{Dong \bgroup \em et al.\egroup
  }{2017}]{dong2017metapath2vec}
Yuxiao Dong, Nitesh~V. Chawla, and Ananthram Swami.
\newblock Metapath2vec: Scalable representation learning for heterogeneous
  networks.
\newblock In {\em Proc.\ of the 23rd ACM SIGKDD International Conference on
  Knowl. Discovery and Data Mining}, page 135–144, 2017.

\bibitem[\protect\citeauthoryear{Douglas}{2011}]{douglas2011weisfeilerlehman}
Brendan~L. Douglas.
\newblock The weisfeiler-lehman method and graph isomorphism testing.
\newblock {\em arXiv preprint}, 2011.
\newblock arXiv:1101.5211.

\bibitem[\protect\citeauthoryear{Ebli \bgroup \em et al.\egroup
  }{2020}]{ebli2020simplicial}
Stefania Ebli, Micha{\"e}l Defferrard, and Gard Spreemann.
\newblock Simplicial neural networks.
\newblock {\em NeurIPS Workshop in Topological Data Analysis and Beyond}, 2020.

\bibitem[\protect\citeauthoryear{Fang \bgroup \em et al.\egroup
  }{2019}]{mhin_metagraph}
Yang Fang, Xiang Zhao, Peixin Huang, Weidong Xiao, and Maarten de~Rijke.
\newblock M-hin: Complex embeddings for heterogeneous information networks via
  metagraphs.
\newblock In {\em Proc.\ of the 42nd International ACM SIGIR Conference on
  Research and Development in Information Retrieval}, SIGIR'19, page 913–916,
  2019.

\bibitem[\protect\citeauthoryear{Fu \bgroup \em et al.\egroup
  }{2020}]{fu2020magnn}
Xinyu Fu, Jiani Zhang, Ziqiao Meng, and Irwin King.
\newblock {MAGNN}: Metapath aggregated graph neural network for heterogeneous
  graph embedding.
\newblock In {\em Proc.\ of The Web Conference 2020}, page 2331–2341, 2020.

\bibitem[\protect\citeauthoryear{Giusti \bgroup \em et al.\egroup
  }{2016}]{brainnetworkanalysis}
Chad Giusti, Robert Ghrist, and Danielle~S. Bassett.
\newblock Two's company, three (or more) is a simplex.
\newblock {\em Journal of Computational Neuroscience}, 41(1):1--14, 2016.

\bibitem[\protect\citeauthoryear{{GUDHI Project}}{2015}]{gudhi}
{GUDHI Project}.
\newblock {\em {GUDHI} User and Reference Manual}.
\newblock {GUDHI Editorial Board}, 2015.

\bibitem[\protect\citeauthoryear{Hamilton \bgroup \em et al.\egroup
  }{2017}]{hamilton2017graphsage}
William~L. Hamilton, Rex Ying, and Jure Leskovec.
\newblock Inductive representation learning on large graphs.
\newblock In {\em Proc.\ of the 31st International Conference on Neural
  Information Processing Systems}, page 1025–1035, 2017.

\bibitem[\protect\citeauthoryear{Hatcher}{2000}]{Hatcher:478079}
Allen Hatcher.
\newblock {\em {Algebraic topology}}.
\newblock Cambridge Univ. Press, Cambridge, 2000.

\bibitem[\protect\citeauthoryear{Horn \bgroup \em et al.\egroup
  }{2022}]{horn2021topological}
Max Horn, Edward~De Brouwer, Michael Moor, Yves Moreau, Bastian Rieck, and
  Karsten Borgwardt.
\newblock Topological graph neural networks.
\newblock {\em International Conference on Learning Representations}, 2022.

\bibitem[\protect\citeauthoryear{Hussein \bgroup \em et al.\egroup
  }{2018}]{just2018}
Rana Hussein, Dingqi Yang, and Philippe Cudr\'{e}-Mauroux.
\newblock Are meta-paths necessary? revisiting heterogeneous graph embeddings.
\newblock In {\em Proc.\ of the 27th ACM International Conference on
  Information and Knowledge Management}, page 437–446, 2018.

\bibitem[\protect\citeauthoryear{Ji \bgroup \em et al.\egroup
  }{2022}]{ji2020signal}
Feng Ji, Giacomo Kahn, and Wee~Peng Tay.
\newblock Signal processing on simplicial complexes with vertex signals.
\newblock {\em IEEE Access}, pages 1--1, 2022.

\bibitem[\protect\citeauthoryear{Kipf and Welling}{2016}]{kipf2016semi}
Thomas~N Kipf and Max Welling.
\newblock Semi-supervised classification with graph convolutional networks.
\newblock {\em International Conference on Learning Representations}, 2016.

\bibitem[\protect\citeauthoryear{Lee \bgroup \em et al.\egroup
  }{2021}]{regather}
See~Hian Lee, Feng Ji, and Wee~Peng Tay.
\newblock Learning on heterogeneous graphs using high-order relations.
\newblock {\em ICASSP 2021 - 2021 IEEE International Conference on Acoustics,
  Speech and Signal Processing (ICASSP)}, pages 3175--3179, 2021.

\bibitem[\protect\citeauthoryear{Roman \bgroup \em et al.\egroup
  }{2015}]{tumoranalysis}
Theodore Roman, Amir Nayyeri, Brittany Fasy, and Russell Schwartz.
\newblock A simplicial complex-based approach to unmixing tumor progression
  data.
\newblock {\em BMC bioinformatics}, 16:254, 12 2015.

\bibitem[\protect\citeauthoryear{Sankar \bgroup \em et al.\egroup
  }{2019}]{meta-gnn-asonam2019}
Aravind Sankar, Xinyang Zhang, and Kevin Chen-Chuan Chang.
\newblock Meta-gnn: Metagraph neural network for semi-supervised learning in
  attributed heterogeneous information networks.
\newblock In {\em Proc.\ of the 2019 IEEE/ACM International Conference on
  Advances in Social Networks Analysis and Mining}, 2019.

\bibitem[\protect\citeauthoryear{Shi \bgroup \em et al.\egroup }{2019}]{HERec}
Chuan Shi, Binbin Hu, Wayne~Xin Zhao, and Philip~S. Yu.
\newblock Heterogeneous information network embedding for recommendation.
\newblock {\em IEEE Trans.\ on Knowledge and Data Engineering},
  31(2):357–370, February 2019.

\bibitem[\protect\citeauthoryear{Veli{\v{c}}kovi{\'c} \bgroup \em et al.\egroup
  }{2017}]{velivckovic2017graph}
Petar Veli{\v{c}}kovi{\'c}, Guillem Cucurull, Arantxa Casanova, Adriana Romero,
  Pietro Lio, and Yoshua Bengio.
\newblock Graph attention networks.
\newblock {\em International Conference on Learning Representations}, 2017.

\bibitem[\protect\citeauthoryear{Wang \bgroup \em et al.\egroup
  }{2019}]{han2019}
Xiao Wang, Houye Ji, Chuan Shi, Bai Wang, Yanfang Ye, Peng Cui, and Philip~S
  Yu.
\newblock Heterogeneous graph attention network.
\newblock In {\em The World Wide Web Conference}, page 2022–2032, 2019.

\bibitem[\protect\citeauthoryear{Yang \bgroup \em et al.\egroup
  }{2018}]{metagraph-spectral2018}
Carl Yang, Yichen Feng, Pan Li, Yu~Shi, and Jiawei Han.
\newblock Meta-graph based {HIN} spectral embedding: methods, analyses and
  insights.
\newblock In {\em ICDM}, 2018.

\bibitem[\protect\citeauthoryear{Yun \bgroup \em et al.\egroup
  }{2019}]{yun2019gtn}
Seongjun Yun, Minbyul Jeong, Raehyun Kim, Jaewoo Kang, and Hyunwoo~J. Kim.
\newblock Graph transformer networks.
\newblock In {\em Proc.\ of the 33rd International Conference on Neural
  Information Processing Systems}, 2019.

\bibitem[\protect\citeauthoryear{Zhang \bgroup \em et al.\egroup
  }{2020}]{mg2vec}
Wentao Zhang, Yuan Fang, Zemin Liu, Min Wu, and Xinming Zhang.
\newblock mg2vec: Learning relationship-preserving heterogeneous graph
  representations via metagraph embedding.
\newblock {\em IEEE Trans.\ on Knowledge and Data Engineering}, 2020.

\end{thebibliography}
\end{document}


\appendix
\section{Dataset Statistics}
We perform node classification task on three heterogeneous benchmark datasets. Characteristics of the datasets are summarised in \cref{table:dataset_stats}. DBLP has three node types (author(A), paper(P) and conference(C)), and the research area of author serve as labels. ACM consists of three node types (paper(P), author(A) and subject(S)), and categories of papers are to be predicted. IMDB has three node types (movie(M), actor(A), and director(D)). The labels to be determined are the genres of the movies. Each node in DBLP, ACM and IMDB has an associated, bag-of-words node feature.

\begin{table}[!htb]
\captionsetup{justification=centering}
\caption{Summary of datasets}
\label{table:dataset_stats}
\centering
\resizebox{\columnwidth}{!}{
\begin{tabular}{cccccc} 
\toprule
\multicolumn{1}{c}{Dataset} & \multicolumn{1}{c}{\# Nodes} & \multicolumn{1}{c}{\# Edges} & {\# Node type} &{\# Classes}& {\# Features} \\ 
\midrule
DBLP                        & 18405                        & 67946                        & 3                &  4            & 334\\
ACM                         & 8994                         & 25922                         & 3               &  3            & 1902\\
IMDB                        & 12772                        & 37288                        & 3                &  3            & 1256\\
\bottomrule
\end{tabular}}
\end{table}

\section{Preliminaries}\label{sec:SC-intro}
In this section, we give a brief overview of simplicial complexes. We refer interested readers to \cite{Hatcher:478079} for more details. Here, we discuss some properties of simplicial complexes and the notion of upper adjacency, which we use in our proposed SGAT model.

\subsection{Simplicial Complex}\label{subsec:SC}
A simplicial complex is a set consisting of vertices, edges and other higher order counterparts, all of which are known as $k$-simplices, $k\geq0$, defined as follows.

\begin{definition}[k-simplex]\label{def:k-simplex}
A standard (unit) $k$-simplex is defined as 
\begin{align}\label{sigmak}
\sigma^{k} = \set*{(\tilde{v}_0,\ldots,\tilde{v}_{k}) \in \mathbb{R}^{k+1}_{\geq 0} \given \sum_{i=0}^{k}{\tilde{v}_i = 1}}.
\end{align}
Any topological space that is homeomorphic to the standard $k$-simplex and thus, share the same topological characteristics is called a $k$-simplex. 
\end{definition}
As an example, let ${v}_0,\ldots,{v}_{k}$ be vertices of a graph embedded in a Euclidean space so that they are affinely independent (i.e., $v_1-v_0, \ldots, v_k-v_0$ are linearly independent). Then the set 
\begin{align*}
\set*{\sum_{i=0}^k \alpha_i v_i \given \alpha_i\geq 0,\ \sum_{i=0}^k\alpha_i =1}
\end{align*}
 is a $k$-simplex. In particular,the vertices $\{v_i\}$ are instances of $\sigma^0$ while the edges $(v_i, v_j)$, $i\ne j$, are instances of $\sigma^1$. Subsequently, it is possible to geometrically interpret 0-simplices $\sigma^0$ as vertices, 1-simplices $\sigma^1$ as edges, 2-simplices $\sigma^2$ as triangles, 3-simplices $\sigma^3$ as tetrahedron and so on. 

A face of a $k$-simplex is a $(k-1)$-simplex that we obtain from \cref{sigmak} by restricting a fixed coordinate in $(\tilde{v}_0,\ldots,\tilde{v}_{k})$ to be zero. Specifically, a face of a $k$-simplex is a set containing elements of the form $(\tilde{v}_0,\ldots,\tilde{v}_{j-1},\tilde{v}_{j+1},\ldots,\tilde{v}_{k})$. A $k$-simplex has $k+1$ faces.

A simplicial complex is a class of topological spaces that encodes higher-order relationships between vertices. The formal definition is as below.
\begin{definition}[Simplicial complex]
A simplicial complex $\chi$ is a finite set of simplices such that the following holds.
\begin{itemize}
  \item Every face of a simplex in $\chi$ is also in $\chi$.
  \item The non-empty intersection of any two simplices $\sigma_1, \sigma_2$ in $\chi$ is a face of both $\sigma_1$ and $\sigma_2$.
\end{itemize}
\end{definition}

It is also possible to produce a concrete geometric object for each simplicial complex $\chi$. The geometric realisation of a simplicial complex is a topological space formed by glueing simplices together along their common faces \cite{ji2020signal}. For instance, a simplicial complex of dimension 1, consists of two kinds of simplices $\sigma^0$ and $\sigma^1$. By glueing $\sigma^1$ with common $\sigma^0$, we obtain a graph in the usual sense. This means that $\chi^0$ is the set of vertices $V$ in the graph and $\chi^1$ is the set of edges $E$ in the graph.

The dimension of $\chi$ is the largest dimension of any simplex in it and is denoted by $K$. As an example, a simplicial complex of dimension $K=2$ must have at least one 2-simplex (a triangle) and not have any $k$-simplices where $k>2$. The set of all simplices in $\chi$ of dimension $k$ is denoted by $\chi^k$ with cardinality $|\chi^k|$. 

\subsection{Simplicial Adjacencies}
Four kinds of adjacencies can be identified between simplices. They are boundary adjacencies, co-boundary adjacencies, lower-adjacencies and upper-adjacencies  \cite{tsp_barbarossa}. In this paper, we utilize only upper-adjacency so that our SGAT model is equivalent to GAT when $K=1$.
We say that two $k$-simplices in $\chi^k$ are upper-adjacent if they are faces of the same $(k+1)$-simplex. The upper adjacency matrix $A^k \in \mathbb{R}^{|\chi^k| \times |\chi^k|}$ of $\chi^k=\set{\sigma^k_1 \ldots \sigma^k_{|\chi^k|}}$ indicates whether pairs of $k$-simplices are upper-adjacent. The upper adjacency matrix of 0-simplices (nodes), $A^0$, is the usual adjacency matrix of a graph.

For example, two nodes are upper-adjacent if they are connected by an edge and two edges are upper-adjacent if they are part of a triangle. The neighborhood of a $k$-simplex, $N(\sigma^k_i)$ is the set of $k$-simplices upper-adjacent to it. We include self-loops to the upper adjacency matrix so that a $k$-simplex is upper-adjacent to itself. We have
\begin{align}
\label{eqn:neighborhood}
N(\sigma^k_i) = \set*{\sigma^k_j \in \set*{\sigma^k_1 \ldots \sigma^k_{|\chi^k|}} \given A^{k}(i,j)=1}.
\end{align}
We also say that $\sigma_j^k \in N(\sigma_i^k)$ is a neighboring simplex to $\sigma_i^k$.

\section{Parameter Sensitivity}
\begin{figure}[!htb]
\centering
	\includegraphics[width=1\columnwidth]{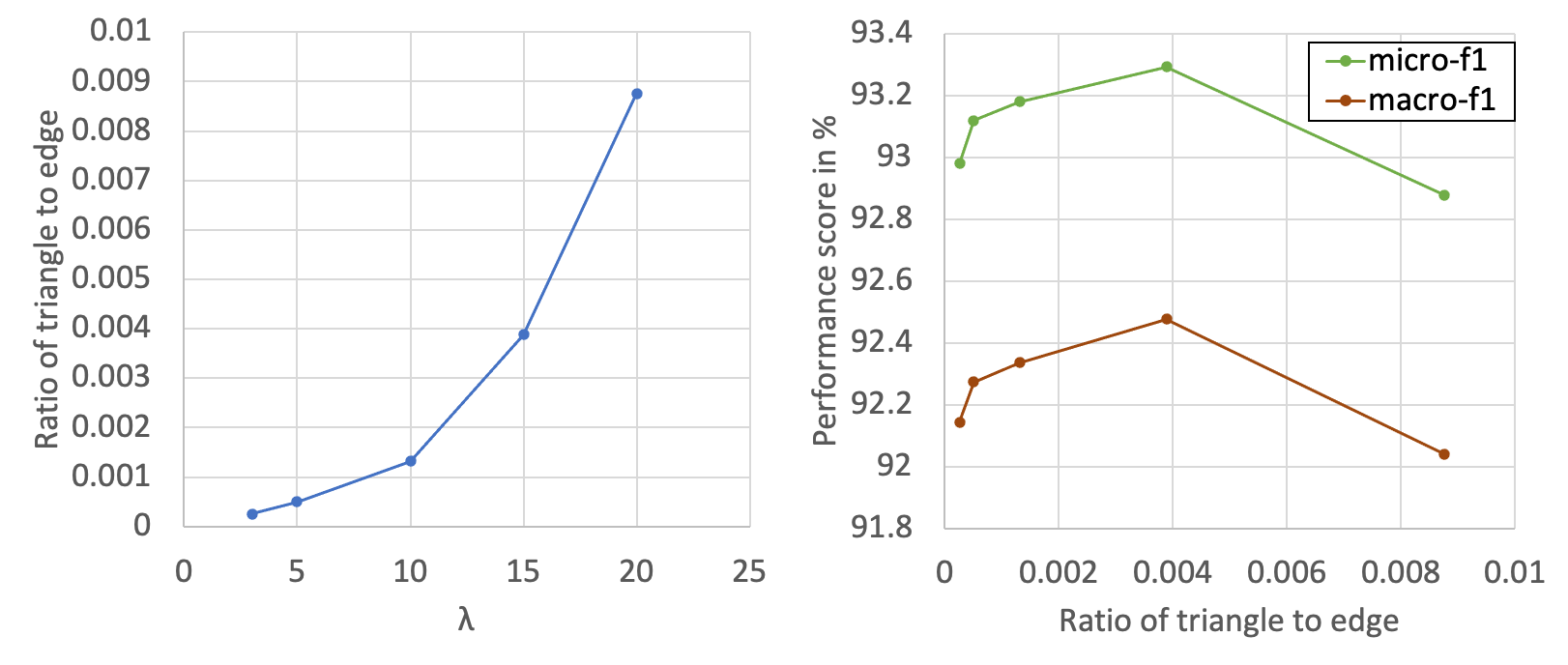}
\caption{$\lambda$ against $\gamma$ (left) and $\gamma$ against performance score (right)}\label{chart:lambda}
\end{figure}

Given that SGAT involves a number of parameters to control the construction of simplices, we further examine how $\lambda$ (the maximum simplex order to construct, including their faces. The simplex order refers to its dimension) and $\epsilon^{k}_{\eta}$ (the minimum number of shared non-target nodes that is $k$ and $\eta$-specific) affect the performance of SGAT on the DBLP dataset. We first keep all the constructed edges by setting $\epsilon^{1}_{1}=\epsilon^{1}_{2}=1$ and measure the ratio of constructed triangles to constructed edges, $\gamma$ as a function of $\lambda$. The Micro-F1 and Macro-F1 scores are then obtained as a function of $\gamma$ as seen in \cref{chart:lambda}. We observe that there is an optimal $\gamma$ where the information included from the amount of simplices considered is most beneficial and does not negatively affect the model's performance. Increasing $\lambda$ beyond the resulting optimal $\gamma$ is likely to introduce noise by including unnecessary messages being passed between the simplices.

We also analyse how $\epsilon^{k}_{\eta}$ affect SGAT's performance by setting $\lambda$ to be fixed at 10 and the default values of $\epsilon^{1}_{1}=3$ and $\epsilon^{1}_{2}=4$. Besides the parameter being tested, the other parameters assume their default values. When $\epsilon^{1}_{1}$ and $\epsilon^{1}_{2}$ are increased, the number of constructed edges decreases. We observe that increasing $\epsilon^{1}_{1}$ improves the performance until the default value of $\epsilon^{1}_{1}=3$ is reached. After which, the performance deteriorates. This phenomenon is similar for $\epsilon^{1}_{2}$. Increasing $\epsilon^{k}_{\eta}$ reduces noise in the model as it removes weaker edges (those that shared fewer nodes).

\begin{figure}[!htb]
\centering
	\includegraphics[width=1\columnwidth]{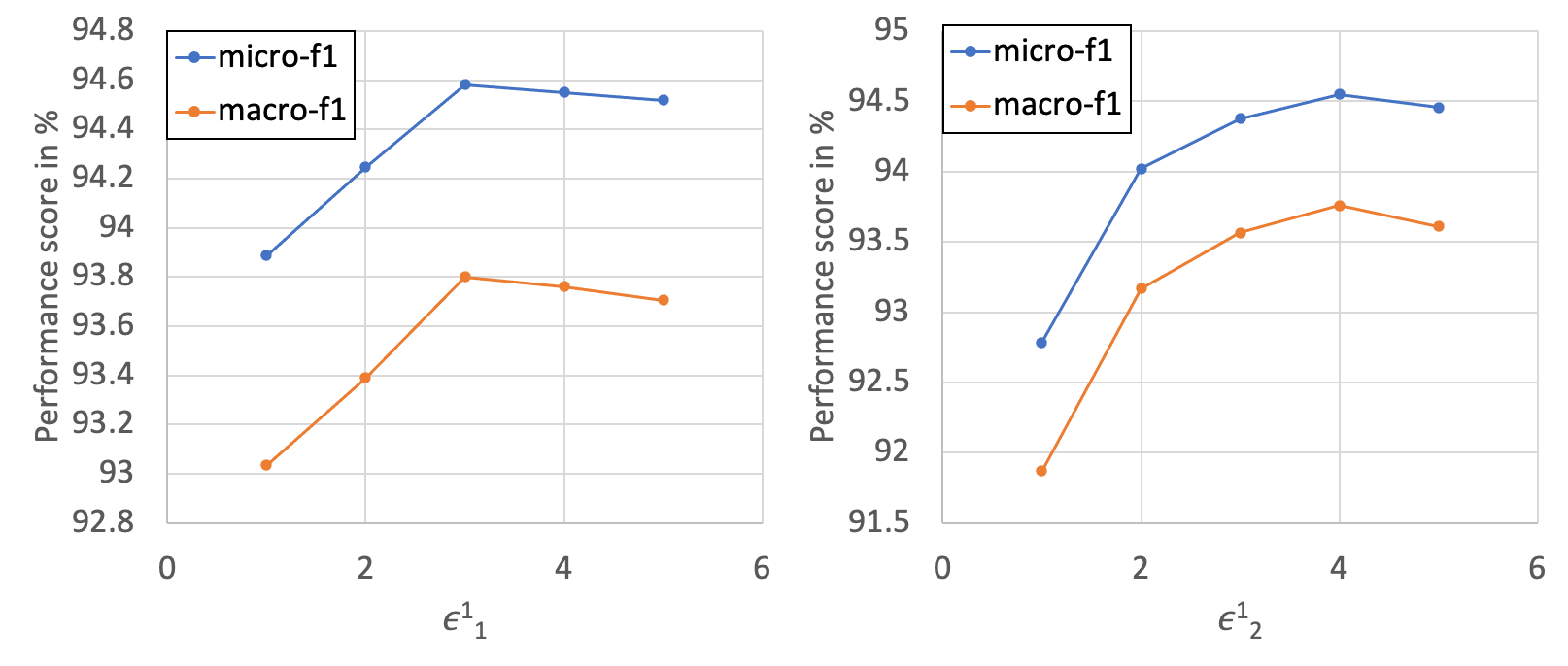}
\caption{$\epsilon^{1}_{1}$ against $\gamma$ (left) and $\epsilon^{1}_{2}$ against $\gamma$ (right)}\label{chart:epsilon}
\end{figure}

Lastly, we examine how different combinations of $\epsilon^{1}_{1}$ and $\epsilon^{1}_{2}$, each chosen in the range of $[1,5]$, affect SGAT's performance. Interestingly, we found that the performance of SGAT fluctuates minimally across different values of these parameters when the ratio of triangles to edges, $\gamma$ is similar. 

\begin{figure}[!htb]
\centering
	\includegraphics[width=1\columnwidth]{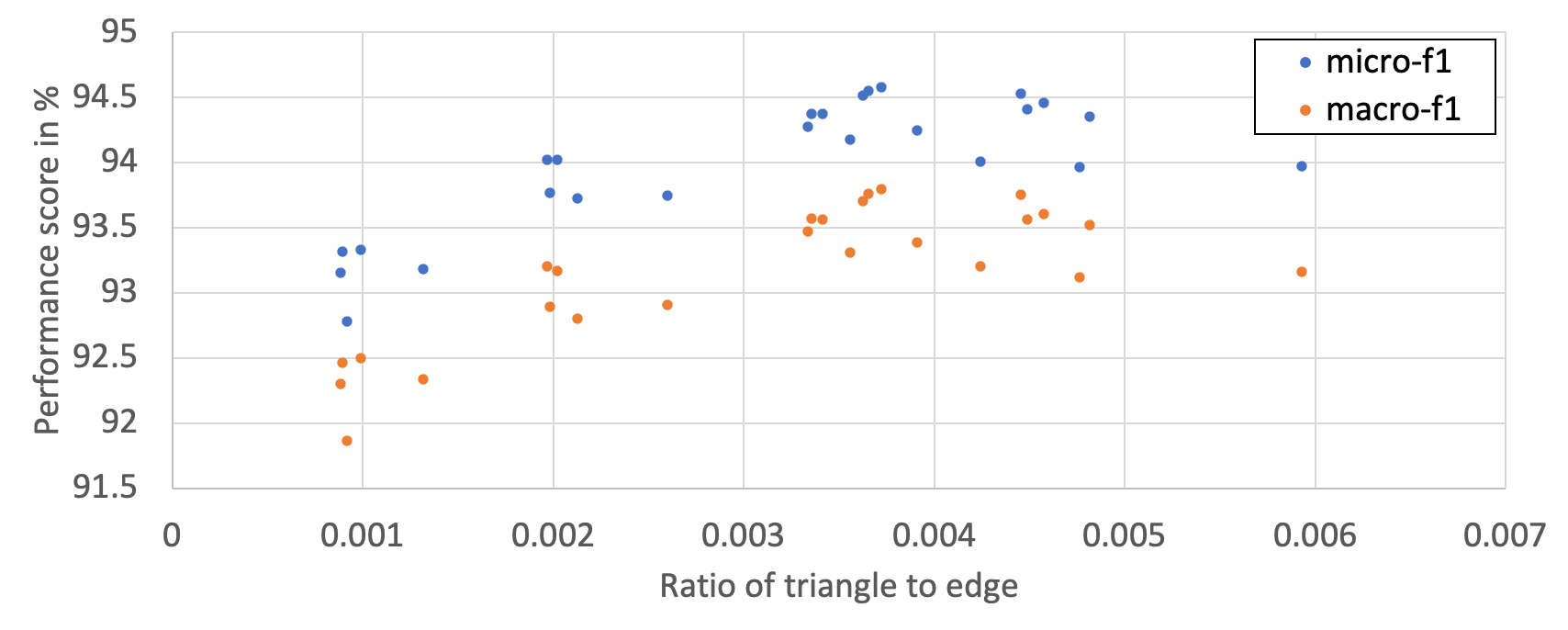}
\caption{$\gamma$ against performance score}\label{chart:combination}
\end{figure}

\section{Time Complexity}\label{sec:timecomplexity}

In addition to evaluating the parameter sensitivity of our model, we compare the time complexity of GTN and SGAT for one layer in each model. GTN is the baseline model that learns the appropriate metapath-based structures in an end-to-end manner. In each graph transformer layer, GTN computes $d$ new metapath specific adjacency matrices using explicit matrix multiplication operation between the $d$ pairs of two softly selected adjacency matrices. The $d$ pairs of selected matrices are obtained by applying $1\times1$ convolution filters on $a$ candidate matrices representing edge types. The time complexity of matrix multiplication operation is known to be $O(|V|^3)$ while the time complexity of $1\times1$ convolution is $O(|V|^2)$. Therefore, the time complexity of a GT layer is $O(da|V|^2 + d|V|^3) = O(d|V|^3)$, where $|V|\gg a$ is the number of nodes.

On the other hand, for our simplicial graph attention layer, we utilize the message passing framework which has a time complexity of $O(|E|)$ if the graph adjacency matrix is sparse, where $|E|$ is the number of edges in the graph \cite{wu2019comprehensive}. To be exact, each simplicial attention layer has $K$ graph attention layers, one for each $k$-simplex ($0\leq k<K$). Hence, the time complexity is given by $O(\|A^0\|+\ldots+\|A^{K-1}\|) = O(K\max_k\|A^k\|)$ where $\|A^k\|$ refers to the number of non-zero entries in $A^k$. Moreover, $P$ heads are employed which makes our simplicial graph attention layer's time complexity to be $O(KP\max_k\|A^k\|)$.

Hence, for sparse graphs, our simplicial graph attention layer has a time complexity upper bound that is quadratic in $|V|$ while the upper bound of the GT layer is cubic in $|V|$. We also measure the training time taken for GTN and SGAT as seen in \cref{table:timetaken}. The experiments are conducted on a server with 32 Intel(R) Xeon(R) Silver 4110 CPU @ 2.10GHz. Due to the large $\epsilon^{1}_{1}$ and $\epsilon^{1}_{2}$ with smaller $\lambda$ for DBLP dataset, the upper adjacencies are less dense. Thus, training on ACM dataset is slower than DBLP despite setting $\eta=2$. Therefore, we can reduce memory consumption and training time while achieving comparable results with small $\lambda$ and increasing $\epsilon^{k}_{\eta}$ as long as $\gamma$ is within the dataset's ideal range.

\begin{table}[!htb]
\caption{Average training time per epoch in seconds}
\label{table:timetaken}
\centering
\begin{tabular}{@{}cccccc@{}}
\toprule
Datasets & GTN    & SGAT \\ \midrule
IMDB     & 33.74  & 3.26   \\
ACM      & 69.57 & 38.16   \\ 
DBLP     & 665.59  & 20.64   \\ \bottomrule
\end{tabular}
\end{table}



     

\bibliographystyle{named}
\bibliography{ijcai22}